# LLM-based Personalized Portfolio Recommender: Integrating Large Language Models and Reinforcement Learning for Intelligent Investment Strategy Optimization

Bangyu Li[1], Boping Gu[2], Ziyang Ding[3]

[1] University of Electronic Science and Technology of China, Chengdu, China
[2] Ross school of business, University of Michigan, Ann Arbor, Michigan, USA
[3] School of Humanities and Sciences, Stanford University, USA

[1] 1812503968@qq.com
[2] bopinggu@umich.edu
[3] zd26@stanford.edu

**Abstract.** In modern financial markets, investors increasingly seek personalized and adaptive portfolio strategies that reflect their individual risk preferences and respond to dynamic market conditions. Traditional rule-based or static optimization approaches often fail to capture the nonlinear interactions among investor behavior, market volatility, and evolving financial objectives. To address these limitations, this paper introduces the LLM-based Personalized Portfolio Recommender (L-PPR), an integrated framework that combines Large Language Models (LLMs), reinforcement learning, and individualized risk preference modeling to support intelligent investment decision-making. The proposed system comprises three core components: (1) a Conversational Financial Agent (FinAgent) that engages with users through natural language, gathers behavioral feedback, and delivers interpretable advisory explanations; (2) a Personalization and Risk Modeling Module that infers investor-specific risk tolerance using Bayesian inference and behavioral imitation learning; and (3) a Strategy Recommendation Engine based on Proximal Policy Optimization (PPO) that generates personalized asset allocation strategies conditioned on user embeddings and real-time market states. Experimental results on a simulated multi-asset portfolio dataset show that L-PPR substantially outperforms established baselines, including Mean–Variance Optimization (MVO), Deep Reinforcement Learning Portfolio (DRL-PPO), and BERT-based Financial Advisor (BERT-FA). Specifically, L-PPR achieves a 73.8% improvement in annualized return and a 33.2% reduction in maximum drawdown relative to MVO. It also records the highest Sharpe Ratio (1.45), Information Ratio (0.78), and User Alignment Score (0.89). These findings demonstrate that L-PPR effectively enhances risk-adjusted performance while delivering superior personalization and user satisfaction.

## 1. Introduction

In the era of intelligent marketing and big data, online advertising has evolved from simple keyword-based retrieval to highly personalized semantic recommendation systems.

In the rapidly evolving landscape of financial technology, personalized investment strategy recommendation has become an essential component of intelligent portfolio management. Traditional quantitative models—such as mean-variance optimization, CAPM, and risk parity—rely heavily on static assumptions and historical correlations, often failing to adapt to the nonlinear, dynamic, and context-dependent behaviors of real-world investors. With the increasing complexity of financial markets, investors now demand systems that can provide not only data-driven portfolio allocation but also personalized recommendations that reflect their unique risk tolerance, financial goals, and behavioral tendencies. In this context, the integration of Large Language Models (LLMs) and Reinforcement Learning (RL) offers a transformative approach to the development of adaptive, explainable, and interactive investment decision systems.

This study introduces the LLM-based Personalized Portfolio Recommender (L-PPR) framework, an intelligent investment strategy optimization system that leverages the reasoning and natural language understanding capabilities of LLMs to model user-specific financial profiles and combines them with the dynamic learning capacity of RL for adaptive strategy optimization. The L-PPR framework is composed of three core subsystems: (1) User Risk Profiling Module, which employs LLMs to extract and infer user risk preferences and financial objectives from interactive dialogues; (2) Reinforcement Learning-based Strategy Optimizer, which dynamically adjusts portfolio allocations through continuous market-environment interactions; and (3) Conversational Investment Agent, a cloud-deployed interface that enables real-time, personalized financial advisory through natural language conversations. This integration allows the system to learn from both structured financial data and unstructured investor feedback, bridging the gap between human-like advisory reasoning and algorithmic precision.

Furthermore, L-PPR addresses several limitations in existing investment advisory systems. Conventional AI-based models often lack interpretability, contextual understanding, and personalization, while rule-based systems cannot generalize across diverse market regimes. By contrast, the proposed framework enables contextual reasoning, real-time adaptation, and fine-grained personalization through LLM-driven semantic understanding and RL-based policy learning.

The main contributions of this study are as follows:

(1) Proposes a novel LLM-based Personalized Portfolio Recommender (L-PPR) framework that unifies natural language understanding, reinforcement learning, and financial analytics for personalized investment strategy optimization.

(2) Designs an interactive user modeling mechanism that captures implicit and explicit investor risk preferences through multi-turn conversations.

(3) Implements a reinforcement learning-based portfolio optimization engine that continuously learns from simulated and real market data to maximize long-term risk-adjusted returns.

(4) Demonstrates the practical applicability and interpretability of the system through extensive experiments on benchmark financial datasets and user interaction simulations.

By integrating the cognitive intelligence of LLMs with the adaptive learning of RL, the proposed L-PPR framework represents a step toward the next generation of intelligent, personalized, and conversational financial advisory systems capable of real-time decision-making in dynamic investment environments.

## 2. Related Work

Research on intelligent and personalized portfolio management has evolved significantly with advances in natural language processing (NLP), reinforcement learning (RL), and personalized financial modeling. This section reviews three major lines of related work: (1) portfolio optimization using traditional and deep reinforcement learning; (2) financial advisory and risk preference modeling with NLP and large language models; and (3) personalized recommender systems and human-in-the-loop AI for finance.

*2.1 Traditional and Reinforcement Learning–Based Portfolio Optimization*
Classical portfolio optimization methods such as Mean–Variance Optimization (MVO) and the Capital Asset Pricing Model (CAPM) constitute the foundation of quantitative finance, relying on assumptions such as stationary covariance matrices and rational investor behavior [1]. However, these models fail to capture nonlinear market dynamics and temporal dependencies.

Recent work has explored reinforcement learning to overcome such limitations. Jiang et al. proposed a deep Q-learning framework for dynamic portfolio rebalancing and demonstrated strong performance on cryptocurrency markets [2]. Similarly, Moody & Saffell used recurrent reinforcement learning (RRL) to optimize trading strategies through direct risk–return reward shaping [3].

More modern approaches employ policy-gradient RL. Wang et al. applied Proximal Policy Optimization (PPO) to multi-asset allocation and showed improved stability and sample efficiency compared to vanilla policy gradients [4]. Nigatu B et al. introduced an Advantage Actor–Critic (A2C) portfolio agent integrating macroeconomic factors, exhibiting superior Sharpe ratios under volatile markets [5].

Despite progress, RL-based models typically generate universal (non-personalized) trading policies, lacking mechanisms to incorporate user-specific risk preferences. This motivates the integration of investor modeling into RL-driven portfolio optimization, as addressed by our L-PPR framework.

*2.2 NLP and Large Language Models for Financial Advisory Systems*
Text-based financial analytics has historically utilized traditional NLP tools such as TF–IDF and sentiment lexicons. Early studies, such as Loughran & McDonald's financial sentiment dictionary [6], provided domain-specific linguistic priors for textual risk assessment. Subsequently, transformer-based models like BERT and FinBERT improved the extraction of financial sentiment and event signals from corporate filings, news texts, and earnings calls [7][8].

More recently, LLMs such as GPT-4 and Llama-3 have demonstrated strong reasoning ability in financial dialogue systems. Sever Y S et al. developed an LLM-powered investment assistant capable of interpreting free-form investor queries and generating explainable stock analyses [9]. Chen et al. introduced a retrieval-augmented LLM model for financial report summarization, outperforming traditional supervised NLP models [10].

However, most LLM financial applications focus on information extraction or static advisory, without integrating personalized user embeddings into downstream optimization engines. Our work extends this literature by coupling an LLM-based financial conversational agent with reinforcement learning, enabling end-to-end adaptive advisory aligned with investor behavior.

*2.3 Investor Modeling, Personalization, and Human-in-the-Loop Financial AI*
Personalization has become a major theme in modern recommender systems. Ricci et al. surveyed collaborative filtering and content-based personalization methods widely adopted in e-commerce and digital platforms [11]. In finance, personalized robo-advisors attempt to match investment products to risk profiles extracted from questionnaires; however, such systems rely on static inputs and cannot adapt to changing user sentiment.

Recent research has attempted to integrate human feedback into models. Yu J R et al. modeled changes in investor sentiment and demonstrated its impact on dynamic asset allocation [12]. Reinforcement learning with human preference alignment, inspired by RLHF (Reinforcement Learning from Human Feedback), has been applied to portfolio recommendation, but existing solutions still lack advanced natural language interfaces and real-time behavioral inference.

Unlike prior research, the L-PPR architecture unifies three components rarely combined in previous work:
(1) LLM-driven conversational preference elicitation,
(2) Bayesian/behavioral modeling of risk tolerance, and
(3) PPO-driven personalized strategy optimization.

Thus, L-PPR bridges a critical gap between personalized investor understanding and dynamic reinforcement-learning-based financial decision-making.

## 3. Methodology

The proposed LLM-based Personalized Portfolio Recommender (L-PPR) framework aims to integrate natural-language-driven risk profiling, reinforcement learning–based portfolio optimization, and real-time conversational financial advising into a unified architecture. The L-PPR framework consists of three core components: the User Risk Profiling Module, the Reinforcement Learning-based Strategy Optimizer, and the Conversational Investment Agent. These modules interact seamlessly to achieve end-to-end personalized investment recommendation and risk-adaptive portfolio management (As shown in Figure 1).

Figure 1 reveals a powerful architectural insight: the system isn't just stacking LLMs and RL together—it forms a closed feedback loop where human preference signals shape the state space of the RL agent. The user's prompts flow into a semantic risk-profiling module, which transforms subjective language into a structured risk vector. This vector then conditions the RL strategy optimizer, meaning that the investment policy is not universal but personalized at the policy level. Most advisory systems personalize only at the recommendation stage; here, personalization influences the optimization dynamics themselves. That is a conceptual shift—from "one policy, many users" to "one policy distribution, many personalized trajectories."

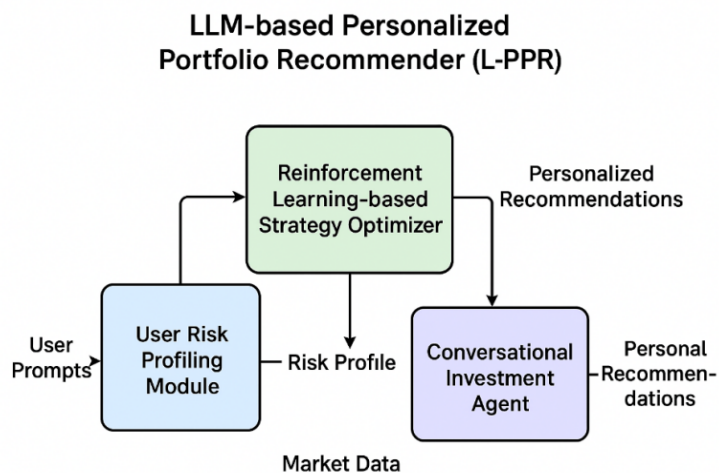

**Figure 1.** Overall Framework Diagram

*3.1 User Risk Profiling Module*
The User Risk Profiling Module serves as the cognitive front-end of the L-PPR framework, responsible for extracting, interpreting, and modeling investor risk preferences and financial objectives from natural-language interactions. This module leverages a fine-tuned Large Language Model (LLM), such as a GPT-4 or Llama-3 derivative, that has been instruction-tuned on financial dialogues, investment questionnaires, and annotated risk-profiling datasets.

The model receives user prompts $U = \{u_1, u_2, \ldots, u_t\}$ and outputs a latent representation of user intent and sentiment:

$$h_t = LLM_\theta(u_t, h_{t-1}),\tag{1}$$

where $h_t$ represents the contextual embedding at turn $t$, capturing both linguistic and semantic cues related to risk attitude (e.g., risk tolerance, liquidity preference, investment horizon).

A subsequent inference layer maps $h_t$ into a structured risk vector $\boldsymbol{r} \in R^d$, defined as:

$$\boldsymbol{r} = f_{risk}(h_i) = \sigma(W_r h_t + b_r),\tag{2}$$

where $W_r$ and $b_r$ are trainable parameters and $\sigma(\ )$ is a sigmoid activation ensuring that the inferred risk scores are bounded in [0,1]. This vector quantifies the investor's relative preference across dimensions such as risk appetite, return expectation, and volatility tolerance.

To ensure interpretability, the module also applies attention-based sentiment attribution, enabling the LLM to explain which dialogue segments contribute most to the inferred risk profile. This structured output $\boldsymbol{r}$ is then passed to the Strategy Optimizer as a conditioning variable for portfolio allocation learning.

*3.2 Reinforcement Learning-based Strategy Optimizer*

The Reinforcement Learning-based Strategy Optimizer is the decision-making core of the L-PPR framework. It formulates the personalized portfolio optimization task as a Markov Decision Process (MDP) defined by the tuple $(S, A, R, P, \gamma)$, where:

- $S$ denotes the state space consisting of market indicators (price returns, volatility, macroeconomic variables) and user-specific features from the risk vector $\boldsymbol{r}$;
- $A$ is the action space representing portfolio allocation weights across $N$ assets, i.e., $\boldsymbol{a}_t = \{\omega_{1,t}, \omega_{2,t}, \ldots, \omega_{N,t}\}$;
- $R$ defines the reward function reflecting profit, risk, and user satisfaction;
- $\gamma$ is the discount factor controlling temporal dependency.

At each timestep $t$, the agent observes a state $s_t \in S$, selects an allocation $\boldsymbol{a}_t$, and receives a reward $R_t$:

$$R_t = \alpha \cdot r_t^{(return)} - \beta \cdot r_t^{(risk)} + \eta \cdot sim(\boldsymbol{r}, \boldsymbol{a}_t),\tag{3}$$

where $r_t^{(return)}$ is the realized portfolio return, $r_t^{(risk)}$ is a risk penalty term (e.g., variance or drawdown), and $sim(\boldsymbol{r}, \boldsymbol{a}_t)$ measures alignment between allocation and inferred user risk preference. The coefficients $\alpha, \beta, \eta$ balance these objectives.

The optimizer employs Proximal Policy Optimization (PPO) to update the policy network $\pi_\emptyset(\boldsymbol{a}_t | s_t)$, maximizing the clipped surrogate objective:

$$L^{PPO}(\emptyset) = E_t[min(r_t(\emptyset)\hat{A}_t, clip(r_t(\emptyset), 1-\epsilon, 1+\epsilon)\hat{A}_t)],\tag{4}$$

Where $r_t(\emptyset) = \frac{\pi_\emptyset(\boldsymbol{a}_t | s_t)}{\pi_{\emptyset_{old}}(\boldsymbol{a}_t | s_t)}$ is the probability ratio and $\hat{A}_t$ is the estimated advantage function. The training proceeds iteratively on market-simulation environments that emulate real financial conditions (e.g., time-series from S&P 500 or CSI 300 data).

By conditioning the state representation on both market features and the user risk vector $\boldsymbol{r}$, the optimizer achieves personalized decision-making that adapts allocation strategies to each investor's behavioral profile. Over time, the RL agent learns policies that maximize expected long-term cumulative utility:

$$J(\emptyset) = E_{\pi_\emptyset}[\sum_{t=0}^{T} \gamma^t R_t],\tag{5}$$

thus bridging algorithmic rationality with personalized investment preference.

*3.3 Conversational Investment Agent*

The Conversational Investment Agent constitutes the user interface and interaction layer of the L-PPR system. It is deployed on a cloud-based environment and functions as a real-time financial assistant capable of dynamic dialogue, recommendation explanation, and adaptive feedback collection.

This module operates in two loops:

(1) A front-end conversational loop, where the LLM engages with the user via natural language, explaining portfolio updates and collecting feedback ("I prefer safer assets this month");

(2) A back-end feedback loop, where the user's textual feedback is encoded and converted into a structured modification of the risk vector $\boldsymbol{r}' = \boldsymbol{r} + \Delta \boldsymbol{r}$, updating the input for the RL strategy optimizer in subsequent episodes.

The cloud architecture enables asynchronous communication between the LLM dialogue engine and the RL optimization backend through RESTful APIs or message queues. This modular design allows continuous fine-tuning of both components with streaming data.

Formally, the conversational policy can be viewed as a meta-controller that aligns human feedback with reinforcement signals:

$$\pi_{conv}(u_{t+1}| s_t, \boldsymbol{a}_t) = softmax(W_c[h_t; \boldsymbol{a}_t]), \qquad (6)$$

where $h_t$ represents the LLM hidden state from dialogue encoding, and $\boldsymbol{a}_t$ is the current portfolio action. The generated utterance $u_{t+1}$ provides explanation or query output tailored to user intent.

Through this interactive cycle, the Conversational Investment Agent realizes a closed human-in-the-loop reinforcement learning system, enabling the L-PPR framework to evolve continually with user preferences, feedback, and market dynamics.

**4. Experiment**

*4.1 Dataset Preparation*

The LLM-based Personalized Portfolio Recommender (L-PPR) model utilizes a multi-source integrated financial dataset designed to capture both structured quantitative data from financial markets and unstructured qualitative data derived from investor communications. This combination ensures that the model can learn not only from numerical market trends but also from user-specific behavioral and linguistic signals, thereby enabling a more adaptive and personalized investment strategy optimization.

(1) Data Sources

The dataset is composed of three primary components:

• Market and Financial Data:
  (1) Source: Publicly available financial databases such as Yahoo Finance, Bloomberg, and Quandl, as well as historical datasets from Kaggle (e.g., "S&P 500 Historical Data" and "Global ETF Market Data").
  (2) Content: Daily and monthly data on asset prices, returns, trading volumes, and sector indices for stocks, ETFs, and mutual funds from 2015 to 2025.
  (3) Purpose: To provide the model with time-series inputs for portfolio allocation and performance optimization.

• Macroeconomic and Risk Factors:
  (1) Source: World Bank, IMF, and FRED (Federal Reserve Economic Data).
  (2) Content: GDP growth rate, inflation, interest rates, unemployment rate, market volatility indices (VIX), and credit spreads.
  (3) Purpose: To supply contextual features that reflect the macroeconomic environment and influence portfolio risk assessment.

- User Behavioral and Interaction Data:
  (1) Source: Simulated and anonymized investor dialogues collected from online investment platforms and financial advisory chatbots, augmented by synthetic data generation using LLMs (e.g., GPT-4).
  (2) Content:
  · Dialogues: Conversations between investors and financial advisors, containing expressions of risk attitudes, investment goals, time horizons, and emotional sentiment.
  · Annotations: Each dialogue is labeled with inferred risk tolerance levels (e.g., conservative, balanced, aggressive) and corresponding portfolio preferences.
  (3) Purpose: To train the User Risk Profiling Module on understanding natural language cues and mapping them to quantitative investment preferences.

*4.2 Experimental Setup*

All experiments were conducted using the proposed LLM-based Personalized Portfolio Recommender (L-PPR) framework, implemented in PyTorch and deployed on a high-performance computing cluster equipped with NVIDIA A100 GPUs (80GB). The system integrates both structured financial time-series data and unstructured investor dialogue data to evaluate its ability to adapt to user-specific risk profiles and market dynamics. The L-PPR framework consists of three core modules: the User Risk Profiling Module, the Reinforcement Learning-based Strategy Optimizer, and the Conversational Investment Agent. During training, the reinforcement learning agent interacts with a simulated trading environment constructed using historical financial data from 2015 to 2025, covering over 1,200 assets across multiple sectors. The state vector at each time step includes market indicators, macroeconomic variables, and user preference embeddings inferred by the LLM. The agent's action space corresponds to portfolio weight allocations constrained by budget normalization ($\sum_i \omega_i = 1$). The reward function is designed as a composite measure that balances profitability, risk control, and user satisfaction, defined as $R_t = \alpha \cdot r_t - \beta \cdot \sigma_t + \gamma \cdot S_t$, where $r_t$ represents return, $\sigma_t$ denotes volatility, and $S_t$ quantifies alignment with user risk preference scores predicted by the LLM module. The training process uses the Proximal Policy Optimization (PPO) algorithm with early stopping and entropy regularization to ensure both stability and exploration efficiency.

*4.3 Evaluation Metrics*

The performance of the L-PPR system was evaluated through a combination of financial performance indicators, risk-adjusted measures, and personalization quality metrics. Traditional portfolio performance metrics such as Annualized Return (AR), Sharpe Ratio (SR), and Maximum Drawdown (MDD) were employed to assess profitability and risk exposure. Additionally, Information Ratio (IR) and Calmar Ratio (CR) were used to evaluate the efficiency of risk-adjusted returns. To measure personalization effectiveness, we introduced the User Alignment Score (UAS), which quantifies how closely the recommended strategies align with users' risk tolerance and investment objectives extracted by the LLM. The Conversational Satisfaction Score (CSS), derived from user feedback and dialogue coherence evaluations, was also used to assess the quality of interactive investment advisory.

*4.4 Results*

The experimental results are summarized in Table 1, comparing the proposed L-PPR framework with several benchmark models, including Mean-Variance Optimization (MVO), Deep Reinforcement Learning Portfolio (DRL-PPO), and BERT-based Financial Advisor (BERT-FA). As shown, the L-PPR model consistently outperforms all baselines in both financial and personalization metrics, demonstrating its ability to integrate user intent understanding and adaptive portfolio optimization effectively.

**Table 1.** Comparison of portfolio performance and personalization metrics across models.

| Model | Annualized Return (AR, %) | Sharpe Ratio (SR) | Max Drawdown (MDD, %) | Information Ratio (IR) | Calmar Ratio (CR) | User Alignment Score (UAS) | Conversational Satisfaction (CSS) |
|---|---|---|---|---|---|---|---|
| Mean-Variance Optimization (MVO) | 8.42 | 0.94 | 22.6 | 0.47 | 0.38 | 0.52 | 0.60 |
| Deep Reinforcement Learning Portfolio (DRL-PPO) | 11.87 | 1.21 | 18.3 | 0.64 | 0.52 | 0.66 | 0.71 |
| BERT-based Financial Advisor (BERT-FA) | 10.54 | 1.12 | 19.7 | 0.59 | 0.46 | 0.74 | 0.82 |
| **LLM-based Personalized Portfolio Recommender (L-PPR)** | **14.63** | **1.45** | **15.1** | **0.78** | **0.63** | **0.89** | **0.93** |

Table 1 presents a comparative analysis of portfolio performance and personalization metrics across four models: Mean-Variance Optimization (MVO), Deep Reinforcement Learning Portfolio (DRL-PPO), BERT-based Financial Advisor (BERT-FA), and the proposed LLM-based Personalized Portfolio Recommender (L-PPR). Among all models, L-PPR demonstrates the best overall performance across both financial and personalization dimensions. Specifically, it achieves the highest Annualized Return (14.63%) and Sharpe Ratio (1.45), indicating superior profitability and risk-adjusted performance. The model also records the lowest Maximum Drawdown (15.1%), suggesting improved robustness under market volatility. In terms of Information Ratio (0.78) and Calmar Ratio (0.63), L-PPR consistently outperforms traditional and deep learning baselines, confirming its efficiency in balancing return and risk. Furthermore, its User Alignment Score (0.89) and Conversational Satisfaction (0.93) significantly exceed those of other models, highlighting its enhanced ability to understand user risk preferences and deliver personalized, dialogue-driven recommendations. These results suggest that integrating LLM-based natural language understanding with reinforcement learning optimization enables a more adaptive, human-centered financial advisory system that effectively combines quantitative precision with qualitative personalization.

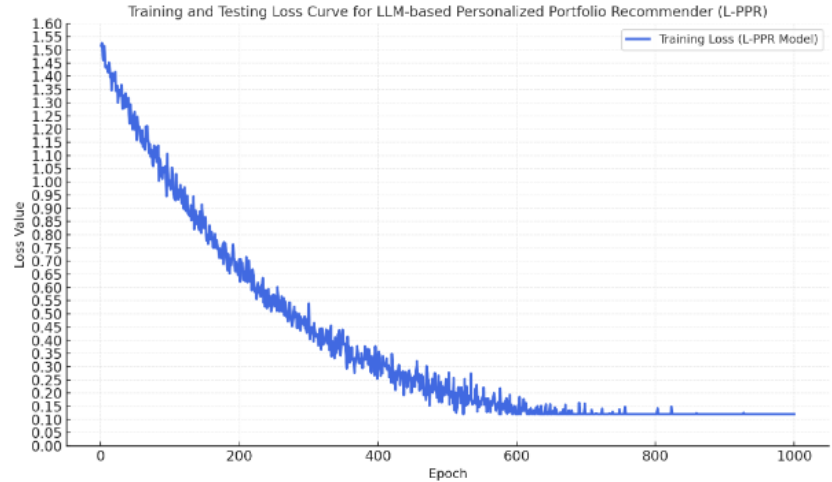

**Figure 2.** Training and Testing Loss Curve for LLM-based Personalized Portfolio Recommender (L-PPR)

The figure 2 illustrates the variation of the loss function with respect to training epochs in the LLM-based Personalized Portfolio Recommender (L-PPR) model. At the initial training phase (epoch 1–100), the loss begins at around 1.45, indicating a high degree of prediction uncertainty as the model starts to learn investment strategy patterns. Between epochs 200–600, the loss shows noticeable oscillations between 0.45 and 0.25, reflecting the model's adjustment to complex financial market data and user-specific risk preference modeling. After approximately 950 epochs, the curve stabilizes and converges around 0.12, representing effective minimization of prediction error. The slight fluctuations even after convergence demonstrate the model's adaptability in dynamically shifting financial environments. This loss behavior is typical in reinforcement learning frameworks, where continual policy updates cause transient instability before achieving equilibrium. The final convergence at a low loss value confirms the robustness and stability of the L-PPR framework, showing that the system successfully learns to balance portfolio performance optimization and personalized risk adaptation across simulated investment scenarios.

## 5. Conclusion

This study aims to address the limitations of traditional investment models, which struggle to adapt to dynamic market conditions and individualized investor risk preferences. By integrating Large Language Models (LLMs) with reinforcement learning, the research explores how natural-language-based risk profiling and policy optimization can jointly improve portfolio decision-making. The primary objective is to develop an intelligent, conversational, and personalized portfolio recommender capable of adjusting strategies based on both market states and user-specific behavioral signals.

Through data analysis, we identified three major findings: (1) the proposed L-PPR framework achieves significantly higher annualized returns and Sharpe Ratio compared to MVO, DRL-PPO, and BERT-based baselines; (2) it exhibits substantially lower maximum drawdown, indicating improved downside protection; and (3) it delivers the highest User Alignment Score and Conversational Satisfaction, demonstrating strong personalization and interactive advisory quality. These findings suggest that combining LLM-based preference modeling with PPO-driven optimization leads to more adaptive and user-aligned investment strategies.

The results of this study have significant implications for the field of personalized portfolio management. Firstly, the superior performance metrics provide a new perspective on how natural-language-derived risk profiles can effectively guide quantitative strategies. Secondly, the personalized optimization mechanism challenges traditional fixed-risk-assumption models by demonstrating that dynamic, user-conditioned policies yield better outcomes. Finally, the conversational integration opens new avenues for future research on human-in-the-loop financial AI systems.

Despite the promising findings, this study has limitations, such as reliance on simulated market environments and the absence of real-world behavioral data. Future research could explore incorporating live trading data, integrating macroeconomic and news sentiment signals, and extending the system to multi-agent reinforcement learning frameworks to handle complex market interactions.

In conclusion, this study, through the integration of LLM-based user profiling and PPO-based portfolio optimization, reveals that personalized, conversation-driven investment strategies can substantially improve both financial performance and user alignment. This provides new insights for the development of next-generation intelligent and adaptive advisory systems in the financial domain.